\documentclass[conference]{IEEEtran}
\IEEEoverridecommandlockouts

\usepackage[table,dvipsnames]{xcolor}
\usepackage{cite}
\usepackage{amsmath,amssymb,amsfonts,dsfont,mathtools}
\usepackage{algorithmic}
\usepackage{graphicx}
\usepackage{textcomp}
\usepackage{multirow,multicol,arydshln}

\usepackage{pifont}
\newcommand{\cmark}{\ding{51}}
\newcommand{\xmark}{\ding{55}}

\makeatletter
\let\MYcaption\@makecaption
\makeatother

\usepackage[font=footnotesize]{subcaption}

\makeatletter
\let\@makecaption\MYcaption
\makeatother

\def\BibTeX{{\rm B\kern-.05em{\sc i\kern-.025em b}\kern-.08em
    T\kern-.1667em\lower.7ex\hbox{E}\kern-.125emX}}

\makeatletter
\newcommand{\linebreakand}{\end{@IEEEauthorhalign} \hfill\mbox{}\par \mbox{}\hfill\begin{@IEEEauthorhalign} }
\makeatother

\usepackage{pdfrender}

\newcommand{\ie}{\textit{i.e.}}
\newcommand{\eg}{\textit{e.g.}}

\begin{document}

\title{DR.WILSS:~
Diffusion-Based Replay for Weakly Supervised Continual Semantic Segmentation \thanks{Code is available at \texttt{https://github.com/LTTM/DRWILSS}}\thanks{This work was partially supported by the European Union under the Italian National Recovery and Resilience Plan (NRRP) Mission 4, Component 2, Investment 1.3, CUP C93C22005250001, partnership on “Telecommunications of the Future” (PE00000001 - program “RESTART”).}}

\author{\IEEEauthorblockN{Leon Arthur Marx}
\IEEEauthorblockA{\textit{Dpt.  Information Engineering} \\
\textit{University of Padova}\\
Padova, Italy \\
leonarthur.marx@dei.unipd.it}
\and
\IEEEauthorblockN{Francesco Barbato}
\IEEEauthorblockA{\textit{Dpt. Information Engineering} \\
\textit{University of Padova}\\
Padova, Italy \\
francesco.barbato@dei.unipd.it
}
\and 
\IEEEauthorblockN{Matteo Caligiuri}
\IEEEauthorblockA{\textit{Dpt.  Information Engineering} \\
\textit{University of Padova}\\
Padova, Italy \\
matteo.caligiuri@dei.unipd.it
}
\and
\IEEEauthorblockN{Pietro Zanuttigh}
\IEEEauthorblockA{\textit{Dpt.  Information Engineering} \\
\textit{University of Padova}\\
Padova, Italy \\
zanuttigh@dei.unipd.it}
}

\maketitle

\begin{abstract}
Weakly supervised class-incremental semantic segmentation (WILSS) aims to train a segmentation model over multiple steps, each introducing new concepts to be learned with only image-level supervision. 
We introduce DR.WILSS, an innovative approach to address catastrophic forgetting in continual learning using diffusion-based generative replay. 
Our framework leverages language clues to guide the diffusion process, employing self-inpainting and regularization techniques to efficiently produce replay data, aiding the learning process. 
By generating high-quality replay data, the information from previously learned classes can be preserved during continual updates, a critical challenge in incremental learning scenarios. 
To further align the statistics of replay data with those of training samples, we apply LoRAs to the generative model. 
Experimental results demonstrate state-of-the-art performance across multiple benchmarks and generative architectures, while avoiding storage of training data and the use of additional resource-demanding tools during training. 
The proposed technique enables an optimal tradeoff between training complexity and inference-time accuracy, making DR.WILSS a promising solution for real-world applications.
\end{abstract}

\begin{IEEEkeywords}
Continual Learning, Semantic Segmentation, Weakly Supervised Learning, Generative Replay
\end{IEEEkeywords}

\section{Introduction and Related Work} \label{sec:introduction}
Semantic Segmentation (SS) refers to pixel-level labeling of images, which allows fine-grained scene understanding in diverse applications, including autonomous driving, robotics, medical imaging, and multimedia understanding. 
Many of these applications require adaptation to dynamic environments with new concepts to be learned, personalization needs, and strict privacy constraints. 
Learning new concepts through simple fine-tuning suffers from \textit{catastrophic forgetting} of already learned ones, while full retraining is computationally expensive and often infeasible due to privacy restrictions or limited access to old data. 
Continual~Learning~(CL) strategies enable efficient learning of new tasks without sacrificing past performance \cite{plop,recall,tikp,mbs}. 
The need for continuous adaptation further emphasizes the cost of annotation. Creating pixel-level labels is labor-intensive, especially in expert-driven domains, which motivates weakly supervised approaches~\mbox{\cite{learningweb,rasp,wish,teddy,wilson}}. These methods rely on cheaper labels, such as image-level ones, making them suited for real-world settings.

Cermelli et al.~\cite{wilson} introduced weakly supervised continual learning for semantic segmentation (WILSS) and tackled it by adding a localizer head that leverages class activation maps~\cite{cam} to generate pseudolabels. Combined with Knowledge Distillation (KD), their framework mitigates forgetting and represents a strong baseline.
However, it cannot match fully supervised performance and does not support single-class increments due to its reliance on negative examples.
\begin{figure}[t]
    \centering
    \includegraphics[width=.9\columnwidth]{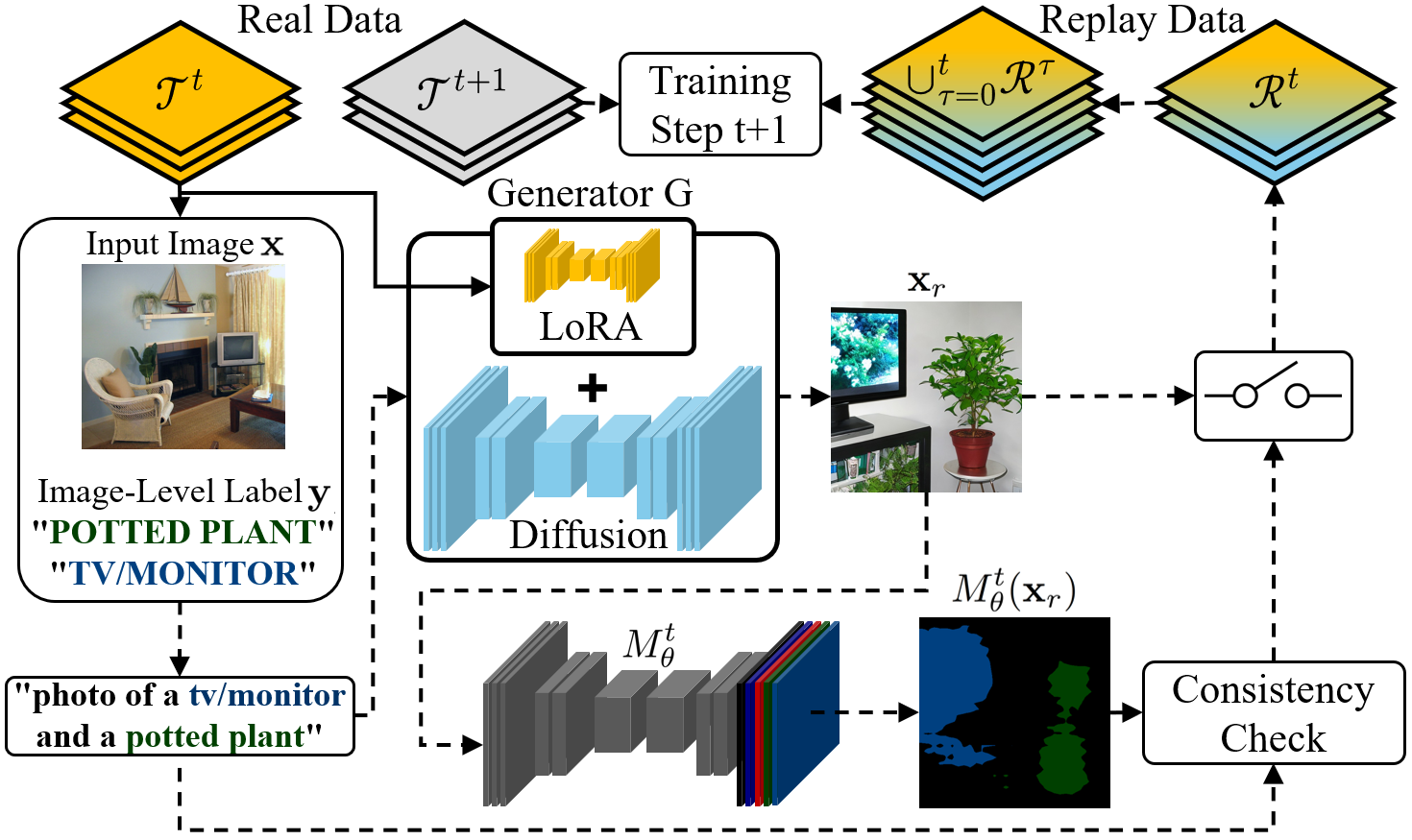}
    \caption{Overview of the DR.WILSS replay strategy: at  each training step the segmentation network is trained with a combination of
    real data and replay samples from the generative module. During training, representative captions are acquired to be used in conjunction with a LoRA module trained on the step data to control the generative model in the next step. A post-generation consistency check ensures that all prompted classes are present.}
    \label{fig:gen_module}
\end{figure}
Several extensions address these limitations. Roy et al.~\cite{rasp} exploit semantic relations to previously seen classes with handcrafted priors and replay images, enabling single-class increments but limiting scalability. Kim and Choe~\cite{wish} similarly use semantic relationships to guide learning, while Si et al.~\cite{teddy} propose a mutual-exclusivity strategy to resolve conflicts between old and new predictions. Liu et al.~\cite{learningweb} extend WILSON \cite{wilson} with web-scraped replay data, but this requires internet access and the availability of suitable public images. Yu et al.~\cite{fmwiss} and Kim et al.~\cite{cwilss} improve replay by assuming access to training data from earlier steps, achieving strong results but narrowing applicability to domains where such data can be stored and reused.
Replay strategies have also been used for continual learning in related fields such as image classification and object detection. 
In \cite{jodelet23}, a diffusion model is used to generate replay-labeled data for class-incremental classification. 
Note that, in the classification task the replay data is labeled, while the dense segmentation labeling needs to be estimated in some way.
This introduces the additional task of producing dense segmentation labels from the weak image-level classification information.
Continual object detection is tackled in \cite{kim24} using custom diffusion models conditioned on bounding box coordinates, thus allowing to generate fully labeled replay data.

In this work, we introduce a diffusion-based generative replay framework for Weakly Supervised Class-Incremental Semantic Segmentation (WILSS). Our method, depicted in Fig.~\ref{fig:gen_module} and denoted DR.WILSS (Diffusion Replay - WILSS), alleviates catastrophic forgetting and reinforces segmentation capabilities by synthesizing examples of previously seen classes.
DR.WILSS is the first approach to: i) exploit diffusion-based replay for
(weakly supervised) continual  semantic segmentation, ii) exploit incrementally tuned LoRAs to enhance the synthetic samples' statistics and to align them with the original training data, iii) introduce novel optimized pseudo-labeling and replay image selection strategies.
These provisions allow our approach to achieve impressive performance on the \mbox{Pascal VOC 2012} and multi-domain COCO-to-VOC (C2V) benchmarks, demonstrating its effectiveness.
Furthermore, unlike prior work, our method requires no additional supervision or handcrafted priors, supports single-class increments, and avoids reliance on external datasets or web images, which may introduce privacy or licensing risks.

\section{Problem Setup} \label{sec:setup}
In the WILSS~\cite{wilson} setting, continual SS is performed without dense supervision labels in the incremental steps, that is, using only image-level classification labels. We assume that training is performed over subsequent steps $t=0,1,\dots$, with corresponding class sets $\mathcal{C}^t$, which we assume to be disjoint for different steps, except for the background class $b$, \ie, $\mathcal{C}^t \cap \mathcal{C}^{t'} \!=\! \{b\}$ for $t \neq t'$. 
At each step $t$, the model receives a task-specific training set $\mathcal{T}^t = \{ (\mathbf{x}_n, \mathbf{y}_n)\}_{n=1}^{N_t}$, consisting of images $\mathbf{x}_n \in \mathbb{R}^{3 \times |\mathcal{I}|}$ and labels $\tilde{\mathbf{y}}_n \in (\mathcal{C}^t)^{|\mathcal{I}|}$ in step $t=0$, and $\mathbf{y}_n \in \{0, 1\}^{|\mathcal{C}^t|}$ in steps $t\geq1$.
Here, $\mathcal{I}$ denotes the set of pixels in a given image. 
That is, we assume pixel-level supervision in step $t=0$, and image-level supervision for all subsequent steps $t\geq1$~\cite{wilson,learningweb,rasp}. 
We denote with {N-M} \cite{mib} the continual learning scenario, where N is the number of classes learned in step $t=0$, while M indicates the classes learned in each subsequent step $t\geq1$. The experiments use two settings: \textit{disjoint}, where future classes do not appear in current-step images, and \textit{overlapped}, which has no such restriction. In both cases, only current-step class labels are provided.

\section{Method} \label{sec:method}
\begin{figure}[t]
    \centering
    \includegraphics[width=.9\columnwidth]{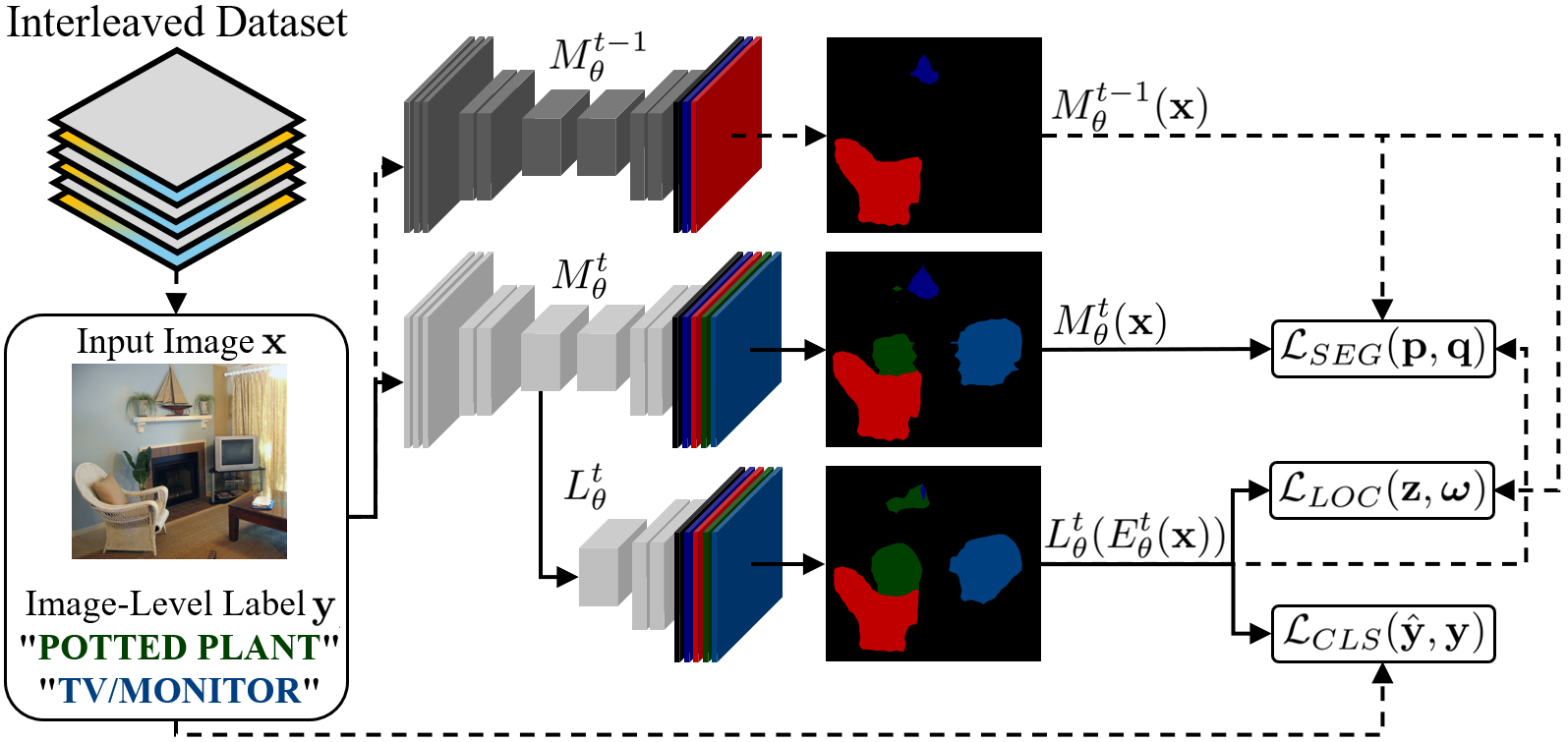}
    \caption{At step $t>0$, the model is trained on new data and replay data of previous classes. The localizer and previous model provide pseudolabels.}
    \label{fig:seg_module}
\end{figure}

Our approach comprises a segmentation and a generative replay module.
The segmentation module (depicted in Fig.~\ref{fig:seg_module}) consists of an encoder-decoder  network $M_{\theta_M} \!=\! D_{\theta_D} \circ E_{\theta_E}$, and a localizer $L_{\theta_L}$. The first is responsible for the pixel-level classification. The second, derived from \cite{wilson}, 
uses the model's latent features to detect the presence of novel classes and provide dense pseudolabels during training. The generative module uses a LoRA-enhanced diffusion model $G$, which provides synthetic replay data to mitigate forgetting.

\subsection{Segmentation Module} \label{ssec:segmentation_module}
To train the segmentation model with only image-level supervision, the localizer needs to correctly detect new classes in a given image and provide accurate pseudolabels.
Given an image-label pair $(\mathbf{x}, \mathbf{y})$, the localizer's pixel-wise predictions are $\textbf{m} \!=\! \text{softmax}(\mathbf{z}) \!\in \! [0, 1]^{|\mathcal{I}| \times |\mathcal{Y}^t|}$, where $\mathbf{z} = L_{\theta_L}(E_{\theta_E}(\mathbf{x}))
$
and $\mathcal{Y}^t = \bigcup_{\tau=0}^t \mathcal{C}^\tau$ the set of all previously seen classes. 
Following \cite{wilson}, they are accumulated over pixel space using normalized Global Weighted Pooling~\cite{ngwp} yielding image-level predictions $\hat{\mathbf{y}}$ used to compute the localizer loss as:
\begin{equation}\label{eq:l_cls}
    \mathcal{L}_{CLS}(\hat{\mathbf{y}}, \mathbf{y}) =  -\frac{1}{|\mathcal{C}^t|}\sum_{c \in \mathcal{C}^t} H(y^c,\hat{y}^c) \;\;,
\end{equation}

\noindent where \mbox{$H(a,b) \!=\! a\log(b)+(1\!-\!a)\log(1\!-\!b)$} is the binary cross-entropy function.
In \cite{wilson}, it is only evaluated on classes of the current step (\ie, $\mathcal{C}^t$), imposing the need for multiple classes to be present in each step to provide the localizer with negative examples, which prevents learning from single-class increments. In our framework, instead, synthetic replay provides the model with negative examples, avoiding this issue.
The localizer is also trained on the previous model's pixel-wise predictions of old classes to improve segmentation accuracy and mitigate forgetting.
Given the previous model's outputs $\boldsymbol{\omega} \!=\! \sigma(M_{\theta_M}^{t-1}(\mathbf{x}))$, the loss is calculated as:
\begin{equation}\label{eq:l_loc}
    \mathcal{L}_{LOC}(\mathbf{z}, \boldsymbol{\omega}) = -\frac{1}{|\mathcal{Y}^{t-1}||\mathcal{I}|} \sum_{i \in \mathcal{I}} \sum_{c \in \mathcal{Y}^{t-1}}H(\omega_i^c,\sigma(z_i^c)) \;\;,
\end{equation}
where $\sigma$ is the sigmoid function. The segmentation network is trained on pseudolabels obtained from a combination of localizer and old model predictions. For new classes, following \cite{wilson}, we use a linear combination of the localizer's hard and soft predictions to compute the pseudolabels, \ie,  defining \mbox{$\tilde{\mathbf{q}} \!=\! \alpha \mathbf{q}^{\text{Hard}} \!+\! (1 \!- \!\alpha) \mathbf{m}$},
where \mbox{$q_i^{\text{Hard},c} \!=\! {\mathds{1}[c \!=\! \text{argmax}_{k \in \mathcal{Y}^t} \hspace{1mm} m_i^k]}$} and $\mathds{1}$ is the indicator function, the pseudolabels are:
\begin{equation}\label{eq:pseudolabel}
    q_i^c =
    \begin{cases}
        \min(\sigma(M_{\theta_M}^{t-1}(\mathbf{x}))_i^c, \tilde{q}_i^c) & \text{if } c = b \\
        \tilde{q}_i^c & \text{if } c \in \mathcal{C}_t \setminus \{b\} \\
        \sigma(M_{\theta_M}^{t-1}(\mathbf{x}))_i^c & \text{else}
    \end{cases} \;\; .
\end{equation}
This means that the background class label is taken as the minimum of the old model and the localizer. 
For new classes, the pseudolabels are calculated from the localizer's prediction, while old class labels are provided by the old model. 
Given the
output $\mathbf{p} = M_{\theta_M}^t(\mathbf{x})$, the  segmentation loss is:
\begin{equation}\label{eq:l_seg}
    \mathcal{L}_{SEG}(\mathbf{p}, \mathbf{q}) =  -\frac{1}{|\mathcal{I}|}\sum_{i \in \mathcal{I}}\sum_{c \in \mathcal{Y}^t} H(q_i^c,\sigma(p_i^c)) \;\; .
\end{equation}

\begin{table*}[t]
    \centering
    \caption{Results in various incremental scenarios, \\
    $^*$ taken from \cite{LIU2026105984}, $^\dagger$ re-implemented for scenarios not reported in \cite{wilson}, \textsuperscript{\textup{(e)}}  method using external data. }
    \label{tab:results}
    \small
    \setlength{\tabcolsep}{2pt}
    \renewcommand{\arraystretch}{1.1}
    \resizebox{\textwidth}{!}{
    \begin{tabular}{cc|c||ccc:ccc|ccc:ccc||ccc:ccc|ccc:ccc}
        \multicolumn{3}{c||}{\multirow{3}{*}{Method}} & \multicolumn{6}{c|}{\textbf{10-5}} & \multicolumn{6}{c||}{\textbf{15-5}} & \multicolumn{6}{c|}{\textbf{10-1}} & \multicolumn{6}{c}{\textbf{15-1}} \\
        \multicolumn{1}{c}{} & \multicolumn{1}{c}{} & \multicolumn{1}{c||}{} & \multicolumn{3}{c}{Disjoint} & \multicolumn{3}{c|}{Overlapped} & \multicolumn{3}{c}{Disjoint} & \multicolumn{3}{c||}{Overlapped} & \multicolumn{3}{c}{Disjoint} & \multicolumn{3}{c|}{Overlapped} & \multicolumn{3}{c}{Disjoint} & \multicolumn{3}{c}{Overlapped} \\
        \multicolumn{1}{c}{} & \multicolumn{1}{c}{} & \multicolumn{1}{c||}{} & 1-10 & 11-20 & All & 1-10 & 11-20 & All & 1-15 & 16-20 & All & 1-15 & 16-20 & All & 1-10 & 11-20 & All & 1-10 & 11-20 & All & 1-15 & 16-20 & All & 1-15 & 16-20 & All \\
        \hline
        \multicolumn{2}{c|}{{\multirow{8}{*}{\vspace*{-2em}\rotatebox{90}{Full Supervision}}}} & ILT$^*$~\cite{iltss} & 53.4 & 48.1 & 50.8 & 55.0 & 44.8 & 49.9 & 31.5 & 25.1 & 29.9 & 69.0 & 46.4 & 63.4 & 14.1 &  0.6 &  7.3 & 16.5 &  1.0 &  8.8 &  6.7 &  1.2 &  5.3 &  5.7 &  1.0 &  4.5 \\
        & & MiB$^*$~\cite{mib} & 54.3 & 47.6 & 51.0 & 55.2 & 49.9 & 52.5 & 71.8 & 43.3 & 64.7 & 75.5 & 49.4 & 69.0 & 14.9 &  9.5 & 12.2 & 15.1 & 14.8 & 14.9 & 46.2 & 12.9 & 37.9 & 35.1 & 13.5 & 29.7 \\
        & & SDR$^*$~\cite{sdr} & 55.5 & 48.2 & 51.9 & 56.9 & 51.3 & 54.1 & 73.5 & 47.3 & 67.0 & 75.4 & 52.6 & 69.7 & 25.5 & 15.7 & 20.6 & 26.3 & 19.7 & 23.0 & 59.2 & 12.9 & 47.6 & 44.7 & 21.8 & 39.0 \\
        & & PLOP~\cite{plop} & - & - & - & - & - & - & 69.1 & 42.8 & 62.5 & 74.1 & 51.7 & 68.5 & - & - & - & 38.4 & 15.5 & 27.0 & 55.1 & 13.7 & 44.7 & 62.8 & 21.1 & 52.4 \\
        & & RECALL$^*$~\cite{recall} & 63.2 & 55.1 & 59.1 & 64.8 & 57.0 & 60.9 & 69.2 & 52.9 & 65.1 & 67.7 & 54.3 & 64.3 & 62.3 & 50.0 & 56.1 & 65.0 & 53.7 & 59.4 & 67.6 & 49.2 & 63.0 & 67.8 & 50.9 & 63.6 \\
        & & TIKP$^*$~\cite{tikp} & - & - & - & 66.2 & 57.9 & 62.1 & - & - & - & 77.4 & 55.5 & 71.9 & - & - & - & 66.7 & 43.5 & 55.1 & - & - & - & 72.1 & 42.3 & 64.6 \\
        & & RECALL+$^*$~\cite{LIU2026105984} & 69.1 & 60.0 & 64.5 & 69.0 & 59.9 & 64.5 & 73.6 & 60.4 & 70.3 & 73.6 & 60.4 & 70.3 & 65.9 & 56.3 & 61.1 & 66.5 & 55.5 & 61.0 & 72.9 & 52.7 & 67.8 & 72.5 & 50.6 & 67.0 \\
        & & Incrementer~\cite{incrementer} & - & - & - & - & - & - & 81.6 & 62.2 & 76.7 & 82.5 & 69.2 & 79.2 & - & - & - & 77.6 & 60.3 & 69.0 & 81.4 & 57.0 & 75.3 & 79.6 & 59.6 & 74.6 \\
        & & MBS~\cite{mbs} & - & - & - & - & - & - & 81.9 & 67.2 & 78.2 & 84.1 & 76.0 & 82.1 & - & - & - & 81.0 & 72.0 & 76.5 & 81.5 & 64.7 & 77.3 & 82.6 & 72.2 & 80.0 \\
        \hline
        \multirow{10}{*}{\vspace*{-1em}\rotatebox{90}{Weak Supervision}} & \multirow{2}{*}{\rotatebox{90}{m.}} & FMWISS~\cite{fmwiss} & - & - & - & - & - & - & 75.9 & 50.8 & 69.6 & 78.4 & 54.5 & 72.4 & - & - & - & - & - & - & - & - & - & - & - & - \\
        & & CWILSS~\cite{cwilss} & - & - & - & - & - & - & 77.3 & 52.2 & 71.0 & 78.8 & 56.9 & 73.3 & - & - & - & - & - & - & - & - & - & - & - & - \\
        \cdashline{2-27}
        & \multirow{5}{*}{\vspace*{-.5em}\rotatebox{90}{no mem.}} & WILSON$^\dagger$~\cite{wilson} & 57.5 & 44.3 & 50.9 & 61.8 & 45.5 & 53.6 & 73.6 & 43.8 & 66.2 & 74.2 & 41.7 & 66.1 &    0 &  0.4 &  0.2 &    0 &  0.3 &  0.2 &    0 &  0.8 &  0.2 &    0 &  0.4 &  0.1 \\
        & & RaSP~\cite{rasp} & \underline{60.5} & \underline{46.8} & \underline{53.6} & 68.8 & 49.1 & 59.0 &  5.9 & 47.5 & 16.3 & 76.2 & 47.0 & 68.9 & \underline{1.3} & \underline{1.0} & \underline{1.1} & \underline{2.0} & \underline{0.7} & \underline{1.4} & \underline{16.2} & \underline{1.8} & \underline{12.6} & \underline{17.7} & \underline{0.9} & \underline{13.5} \\
        & & WISH~\cite{wish} & - & - & - & - & - & - & 74.4 & 45.1 & 67.1 & 75.3 & 45.4 & 67.8 & - & - & - & - & - & - & - & - & - & - & - & - \\
        & & Teddy~\cite{teddy} & - & - & - & \underline{68.9} & \underline{51.7} & \underline{60.3} & \underline{74.5} & \underline{48.1} & \underline{67.9} & \underline{77.6} & \textbf{51.4} & \underline{71.0} & - & - & - & - & - & - & - & - & - & - & - & - \\
        & & \textbf{Ours} & \textbf{66.5} & \textbf{49.6} & \textbf{58.1} & \textbf{71.8} & \textbf{53.1} & \textbf{62.4} & \textbf{77.5} & \textbf{50.0} & \textbf{70.6} & \textbf{78.7} & \underline{49.2} & \textbf{71.4} & \textbf{40.8} & \textbf{22.6} & \textbf{31.7} & \textbf{47.7} & \textbf{24.4} & \textbf{36.1} & \textbf{71.8} & \textbf{25.8} & \textbf{60.3} & \textbf{71.9} & \textbf{25.9} & \textbf{60.4} \\
        \cdashline{2-27}
        &\multirow{2}{*}{\vspace*{-.5em}\rotatebox{90}{ext.}} & Web-WILSS\textsuperscript{(e)}~\cite{learningweb} & - & - & - & - & - & - & 77.1 & 49.1 & 70.1 & 78.2 & 54.9 & 72.4 & - & - & - & 63.9 & 39.2 & 51.5 & - & - & - & 73.6 & 36.2 & 64.2 \\
        & & RaSP $+M_{ext}$\textsuperscript{(e)}~\cite{rasp} & - & - & - & - & - & - & - & - & - & - & - & - & 61.7 & 37.4 & 49.5 & 66.8 & 39.1 & 53.0 & 74.7 & 35.8 & 65.0 & 75.7 & 35.2 & 65.6 \\
        \cline{2-27}
        & & VOC-Replay (Reference) & 68.8 & 52.0 & 60.4 & 73.1 & 56.1 & 64.6 & 77.2 & 53.1 & 71.2 & 78.9 & 55.2 & 73.0 & 63.7 & 51.7 & 57.7 & 66.8 & 57.0 & 61.9 & 74.8 & 47.1 & 67.9 & 77.3 & 48.4 & 70.1 \\
    \end{tabular}}
\end{table*}

\noindent We also apply a local consistency loss $\mathcal{L}_{LC}$, ensuring the same labels for neighboring and similar pixels, and a knowledge distillation loss $\mathcal{L}_{KD}$ aligning the new model's features with the old model ones. For details, refer to \cite{wilson}. 

\subsection{Generative Replay Module} \label{ssec:generative_module}
We propose the idea of using generative models to produce replay data (the experiments use Stable~Diffusion~2.1~\cite{latent_diffusion} and~XL~\cite{sdxl}). The generator performs a mapping $G: \boldsymbol{\kappa} \mapsto \mathbf{x}_r$ from a text prompt $\boldsymbol{\kappa}$ to a synthetic image $\mathbf{x}_r$. We argue that using synthetic samples for replay is a more viable solution compared to storing data or exploiting web images. Generated samples are easier to adapt to different scenarios by changing the prompt or applying LoRA to fine-tune the models. Our approach exploits the second strategy: it finetunes LoRA modules during training and uses them in later steps to align the statistics of replay data to those of training samples.

\textbf{Data Generation:} After each training step $t$, before moving to the next, we generate a corresponding replay caption set $\mathcal{R}^t$ as follows: for each pair $(\mathbf{x}, \mathbf{y}) \in \mathcal{T}^t$, we create a caption $\boldsymbol{\kappa}(\mathbf{y})$ containing all classes present in $\mathbf{y}$. To avoid word ordering biases in diffusion models~\cite{chen2024cat}, we shuffle the order of the classes that should be present in the images. This results in captions of the form \textit{``photo of a motorbike, a person and a car''} (when an image has classes \textit{car, motorbike, person}). Note that we store only captions; there is no visual data storage.

\textbf{Enhanced Generation with LoRAs:}
We train our LoRA modules on a subset of representative samples from the training set.
More specifically, we use \mbox{CLIP ViT-B/32} (vision) to extract descriptors (containing semantic, stylistic, and scene-layout information) from the Pascal samples, before aggregating them per class to obtain prototypes. Note that these are the descriptors used in the \textit{\mbox{CLIP-I}} GenAI alignment metric~\cite{ruiz2023dreambooth}.
The prototypes are used to select the top-k (k=10) images most representative of each class (\ie, those closest to each prototype).
A different LoRA is employed on each incremental step, meaning that most adapters will be trained on multiple classes at the same time. Once an adapter is trained, it will be used in the following steps to enhance the generation of synthetic samples (keeping the generator configuration fixed).

\textbf{Pseudolabeling:} In step $t+1$, we generate the replay image $\mathbf{x}_r = G(\boldsymbol{\kappa}(\mathbf{y}))$ and employ the updated model $M_{\theta_M}^t$ to generate an image-level pseudolabel containing all classes assigned to at least one pixel in $\mathbf{x}_r$ by the current model, \ie, \mbox{$(\hat{\mathbf{y}}_r)_c = \mathds{1}[\exists i: \text{argmax}_k(M_{\theta_M}^t(\mathbf{x}_r))^k_i = c]$}.
During training, pixel-level pseudolabels are obtained using  Eq.~\ref{eq:pseudolabel}.

\textbf{Image Evaluation:} To ensure that generated images have the expected content, we introduce a consistency check, comparing the model prediction $\hat{\mathbf{y}}_r$ with the expected label set in $\mathbf{y}$. More in detail, we check whether all classes present in $\mathbf{y}$ are detected in the image, \ie, $\{c \in \mathcal{C}^t | \mathbf{y}_c = 1\} \setminus \{c \in \mathcal{C}^t | (\hat{\mathbf{y}}_r)_c = 1\} = \emptyset$. If $\hat{\mathbf{y}}_r$ does not fulfill this criterion, the generation process is repeated with a different seed up to 4 times, after which we resort to keeping the last generated image containing only part of the classes. 
This reduces the number of samples with some missing classes, \eg, on Pascal VOC from 581 (over 9568) to 81 in the 15-5 \textit{overlapped} setting and from 2799 (over 22253) to 1112 in the C2V setting. 
Note that the compute cost of this step decreases with the number of iterations.

\textbf{Mixed Training:} During step $t\!+\!1$, the segmentation module is trained on a joint dataset \mbox{$\mathcal{T}^{t+1} \cup \bigcup_{\tau=0}^t \{G(\kappa), \forall\kappa \in \mathcal{R}^\tau\}$} consisting of the real data for the current step and the cumulative synthetic replay set containing all previously seen classes. We enforce a fixed fraction $\rho=0.5$ of generated images per batch (see Sec. \ref{ssec:ablation_study}), and enforce that the model sees each real training sample once every epoch. If the replay caption set is larger than the current training set, we choose a random subset for each epoch. Conversely, if the set is smaller, we extend the replay set by randomly repeating selected captions.

\section{Implementation Details} \label{ssec:implementation_details}
\begin{figure}[t]
    \centering
    \begin{subfigure}{.9\linewidth}
        \centering
        \begin{subfigure}{.05\textwidth}
            \centering
            \rotatebox{90}{~\small{VOC Target}}
        \end{subfigure}
        \begin{subfigure}{.22\textwidth}
            \centering
            \caption*{Bird}
            \includegraphics[width=\textwidth]{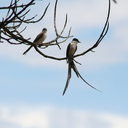}
        \end{subfigure}
        \begin{subfigure}{.22\textwidth}
            \centering
            \caption*{Bottle}
            \includegraphics[width=\textwidth]{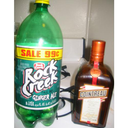}
        \end{subfigure}
        \begin{subfigure}{.22\textwidth}
            \centering
            \caption*{Car}
            \includegraphics[width=\textwidth]{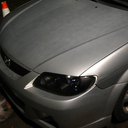}
        \end{subfigure}
        \begin{subfigure}{.22\textwidth}
            \centering
            \caption*{Chair}
            \includegraphics[width=\textwidth]{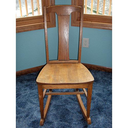}
        \end{subfigure}
    \end{subfigure}
    
    \begin{subfigure}{.9\linewidth}
        \vspace{.3em}
        \centering
        \begin{subfigure}{.05\textwidth}
            \centering
            \mbox{\hspace{-.75em}\rotatebox{90}{~~\small{\parbox{5em}{Zero-Shot (rank = 0)}}}}
        \end{subfigure}
        \begin{subfigure}{.22\textwidth}
            \centering
            \includegraphics[width=\textwidth]{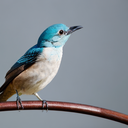}
        \end{subfigure}
        \begin{subfigure}{.22\textwidth}
            \centering
            \includegraphics[width=\textwidth]{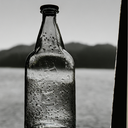}
        \end{subfigure}
        \begin{subfigure}{.22\textwidth}
            \centering
            \includegraphics[width=\textwidth]{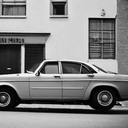}
        \end{subfigure}
        \begin{subfigure}{.22\textwidth}
            \centering
            \includegraphics[width=\textwidth]{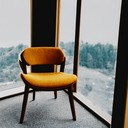}
        \end{subfigure}
    \end{subfigure}
    
    \begin{subfigure}{.9\linewidth}
        \vspace{.25em}
        \centering
        \begin{subfigure}{.05\textwidth}
            \centering
            \rotatebox{90}{~~~\small{rank = 32}~}
        \end{subfigure}
        \begin{subfigure}{.22\textwidth}
            \centering
            \includegraphics[width=\textwidth]{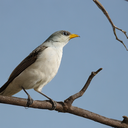}
        \end{subfigure}
        \begin{subfigure}{.22\textwidth}
            \centering
            \includegraphics[width=\textwidth]{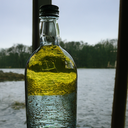}
        \end{subfigure}
        \begin{subfigure}{.22\textwidth}
            \centering
            \includegraphics[width=\textwidth]{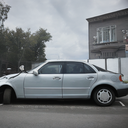}
        \end{subfigure}
        \begin{subfigure}{.22\textwidth}
            \centering
            \includegraphics[width=\textwidth]{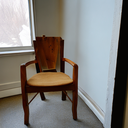}
        \end{subfigure}
    \end{subfigure}

    \begin{subfigure}{.9\linewidth}
        \vspace{.25em}
        \centering
        \begin{subfigure}{.05\textwidth}
            \centering
            \rotatebox{90}{~~~\small{rank = 64}~}
        \end{subfigure}
        \begin{subfigure}{.22\textwidth}
            \centering
            \includegraphics[width=\textwidth]{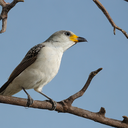}
        \end{subfigure}
        \begin{subfigure}{.22\textwidth}
            \centering
            \includegraphics[width=\textwidth]{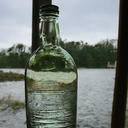}
        \end{subfigure}
        \begin{subfigure}{.22\textwidth}
            \centering
            \includegraphics[width=\textwidth]{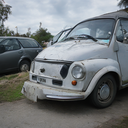}
        \end{subfigure}
        \begin{subfigure}{.22\textwidth}
            \centering
            \includegraphics[width=\textwidth]{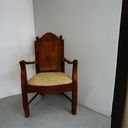}
        \end{subfigure}
    \end{subfigure}

    \begin{subfigure}{.9\linewidth}
        \vspace{.25em}
        \centering
        \begin{subfigure}{.05\textwidth}
            \centering
            \rotatebox{90}{~~\small{rank = 128}}
        \end{subfigure}
        \begin{subfigure}{.22\textwidth}
            \centering
            \includegraphics[width=\textwidth]{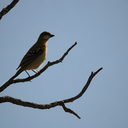}
        \end{subfigure}
        \begin{subfigure}{.22\textwidth}
            \centering
            \includegraphics[width=\textwidth]{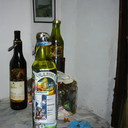}
        \end{subfigure}
        \begin{subfigure}{.22\textwidth}
            \centering
            \includegraphics[width=\textwidth]{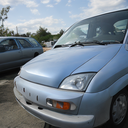}
        \end{subfigure}
        \begin{subfigure}{.22\textwidth}
            \centering
            \includegraphics[width=\textwidth]{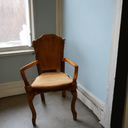}
        \end{subfigure}
    \end{subfigure}
    \caption{Qualitative comparison between different LoRA ranks.}
    \label{fig:lora_rank}
\end{figure}

We evaluate our approach on the commonly used augmented Pascal VOC 2012 benchmark \cite{wilson}, as well as the C2V setting introduced in \cite{wilson}. All experiments were performed with the random seed fixed to 42 (following~\cite{wilson}) on a single Titan RTX GPU. 
We employ the same experimental setup as our \mbox{competitors~\cite{wilson,rasp,wish,teddy,learningweb}}, \ie, a ResNet-101 backbone (stride 16) for Pascal VOC and a Wide-ResNet-38 backbone (stride 8) for C2V, both pretrained on ImageNet-1k, together with a DeepLabv3+ decoder head. In the incremental steps, we optimize the model for 40 (Pascal VOC) or 30 (C2V) epochs using SGD. The batch size is 24 (12 new and 12 replay), momentum 0.9, and weight decay $10^{-4}$. The learning rate is $10^{-3}$ for the backbone and $10^{-2}$ for the localizer and segmentation head. In each step, we freeze all models except the localizer for $5$ epochs as a warm-up phase. The parameter $\alpha$ is set to $0.5$ for Pascal VOC and $0.9$ for C2V. 
We use the standard generation pipeline for both diffusion models considered, in particular, $512\times 512$ px resolution with a DDIM scheduler for the SD2.1 generator (7s per image in fp32), and $1024 \times 1024$ px (downscaled to $512\times 512$ px for training) with a Discrete Euler Scheduler for the SDXL generator (15s per image in fp16). 
The rank of the LoRAs is 128; refer to Figure~\ref{fig:lora_rank} for qualitative comparisons of output images across ranks. Rank 128 consistently produces images that are more realistic and closer to the Pascal VOC samples; we avoid bigger dimensions as they tend to cause overfitting 
issues~\cite{shenaj2026layerscreatedequaladaptive}.

\section{Experimental Results} \label{ssec:quantitative_analysis}
We report the mIoU for the 10-5, 15-5, 10-1, and 15-1 tasks under both \textit{disjoint} and \textit{overlapped} protocols in Tab.~\ref{tab:results}.
Following the standard evaluation protocol, we compute the final mIoU as the mean on all classes excluding background.
Note that including the background typically increases mIoU by $1$–$2\%$. We compare against state-of-the-art memory-free WILSS methods and an upper bound obtained using real Pascal samples for replay data. For completeness, the table also reports results of fully supervised methods and approaches with external memory; these can not be considered fair competitors, as they operate in different (easier) settings.

In the (two-step) 10-5 scenario, our approach surpasses by significant margins the closest weakly supervised competitor in both \textit{disjoint} and \textit{overlapped} protocols, reaching a mIoU of $58.1\%$ ($+4.5\%$) and $62.4\%$ ($+2.1\%$), respectively. This impressive performance is reflected in a very close match to recent fully-supervised methods (\eg, RECALL+~\cite{LIU2026105984}), and in a small drop with respect to the upper-bound reference. 

In the (one-step) 15-5 scenario, our method achieves SOTA performance: being the best in both the \textit{disjoint} and \textit{overlapped} protocols, outperforming competitors by $2.7\%$ in \textit{disjoint} and $0.4\%$ in \textit{overlapped}.
Regardless of the protocol, our approach enjoys a gain of more than $4\%$ from \cite{wilson}, which is our starting codebase.
Our approach achieves the strongest results on the initial step classes, reducing the gap to the reference to $0.2\%$ in \textit{overlapped} and surpassing it by $0.3\%$ in \textit{disjoint}, thus demonstrating that generative replay effectively mitigates catastrophic forgetting. In the incremental step, our method ranks first in \textit{disjoint} and second in \textit{overlapped}, confirming that the preservation of old knowledge does not cause significant issues in learning new classes.
In this setting, our approach also surpasses the external-data method of \cite{learningweb} in \textit{disjoint}, while lagging by $1\%$ in \textit{overlapped}. This result confirms the validity of our approach, as the performance is comparable even with the lack of external web-based data and additional captioning models, which reduce applicability.

\begin{figure}[t]
    \centering
    \includegraphics[width=.9\linewidth]{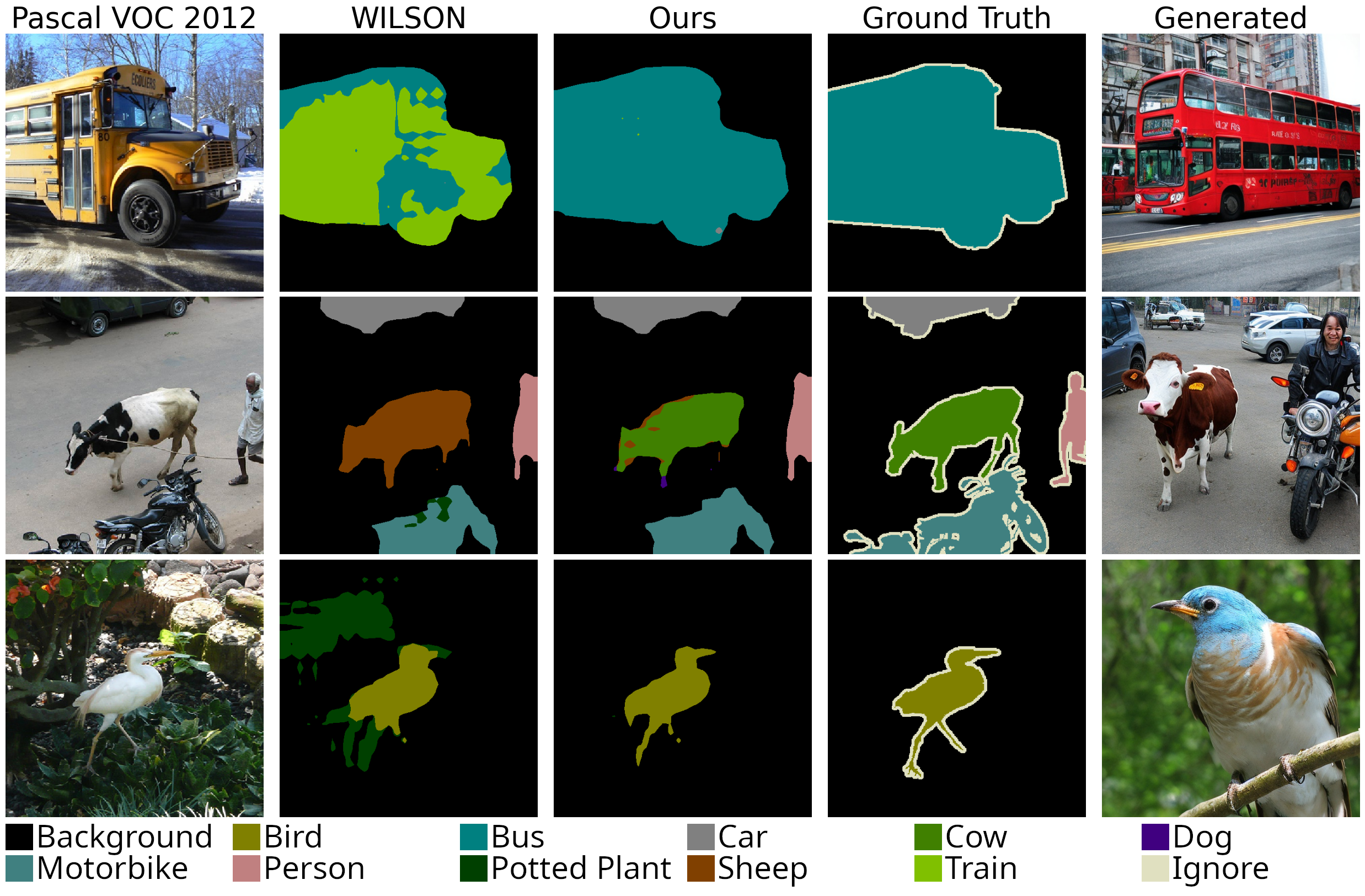}
    \caption{Qualitative results for the \textit{overlapped} 15-5 scenario.}
    \label{fig:qualitative}
\end{figure}

For the 10-1 and 15-1 scenarios, our method effectively learns from single-class incremental steps, reaching $60.3\%$ (\textit{disjoint}) and $60.4\%$ (\textit{overlapped}) in the 15-1 setting, while in the more difficult 10-1 setting, the scores are $31.7\%$ and $36.1\%$, respectively.
The only data-free approach able to tackle this setting with single-class steps is \cite{rasp}, which we outperform by over $30\%$ across all protocols. Approaches using external replay data \cite{learningweb}, and the variants of \cite{rasp} with external memory or stored Pascal samples \cite{fmwiss,cwilss} achieve higher scores in the incremental steps, but they rely on external datasets and labeling pipelines or introduce storage and privacy concerns, thus are not directly comparable.
Our method, instead, is fully self-contained, does not require external data sources or storage, and avoids issues tied to web-scraped data (such as licensing, privacy, adversarial risks, or unavailability in specialized domains) while requiring no manual- or foundation-model-based supervision.
Moreover, we note that generative replay is complementary to replay-free methods: combining it with other continual learning strategies may enhance performance.
Fig.~\ref{fig:qualitative} shows qualitative examples comparing our approach with the closest WILSS competitor WILSON~\cite{wilson}. WILSON confuses previously learned classes with visually similar later ones (\eg, ``bus'' and ``train''  or  ``cow'' and ``sheep''), while our method preserves the acquired knowledge without getting confused by newer, similar classes.
Furthermore, we note how \cite{wilson} fails to resolve overlapping objects (\eg, ``bird'' and ``potted plant'') while our method faithfully resolves even those complex regions. As a final note, we remark how our adapted generation pipeline is able to produce artificial replay data that resembles the real data both in content and style.

\begin{table}[t]
    \caption{Results of the COCO-to-VOC task.}
    \label{tab:cocovoc}
    \centering
    \small
    \renewcommand{\arraystretch}{.9}
    \begin{tabular}{c|ccc|c}
        \multicolumn{1}{c|}{\multirow{2}{*}{Method}} & \multicolumn{3}{c}{COCO} & VOC \\
        & 1-60 & 61-80 & All & 61-80 \\
        \hline
        FMWISS~\cite{fmwiss} & 39.9 & 44.7 & 41.1 & 63.6 \\
        CWILSS~\cite{cwilss} & 38.2 & 45.0 & 39.9 & 64.0 \\
        \hdashline
        WILSON~\cite{wilson} & 39.8 & 41.0 & 40.1 & \underline{55.7} \\
        RaSP~\cite{rasp} & \textbf{41.1} & 40.7 & \textbf{41.0} & 53.6 \\
        WISH~\cite{wish} & 40.2 & \textbf{43.0} & {40.9} & \textbf{58.0} \\
        Teddy~\cite{teddy} & {40.6} & \underline{41.8} & {40.9} & - \\
        \textbf{Ours} & {40.6} & 40.6 & 40.6 & 55.5 \\
       \textbf{Ours - No LoRA} & \underline{40.8} & 41.5 & \textbf{41.0} & \underline{55.7} \\
        \hdashline
        Web-WILSS\textsuperscript{(e)}~\cite{learningweb}   & 40.7 & 33.0 & 38.8 & 56.7 \\
        \hline
        COCO-Replay (Reference) & 42.0 & 40.3 & 41.6 & 53.6 \\
    \end{tabular}
\end{table}

\begin{table*}[t]
    \centering
    \scriptsize
    \setlength{\tabcolsep}{2pt}
    \renewcommand{\arraystretch}{1.1}
    \begin{minipage}[t]{0.295\textwidth}
        \centering
        \caption{Replay data ablation study.}
        \resizebox{\linewidth}{!}{
        \begin{tabular}{c|ccc|ccc}
            \label{tab:abl_rho}
            \multirow{2}{*}{$\rho$}
            & \multicolumn{3}{c|}{15-5 (ov.)}
            & \multicolumn{3}{c}{15-1 (ov.)} \\
             & 1-15 & 16-20 & All & 1-15 & 16-20 & All \\
            \hline
            0.25 & 78.3 & \underline{48.8} & 70.9 & 70.8 & 25.6 & 59.5 \\
            0.5  & \underline{78.7} & \textbf{49.2} & \textbf{71.4} & \textbf{71.9} & \textbf{25.9} & \textbf{60.4} \\
            0.75 & \textbf{78.9} & 48.7 & \underline{71.3} & \underline{71.0} & \underline{25.8} & \underline{59.7} \\
        \end{tabular}
        }
    \end{minipage}\hfill
    \begin{minipage}[t]{0.4\textwidth}
        \centering
        \caption{Generation model ablation study.}
        \renewcommand{\arraystretch}{1.}
        \resizebox{\linewidth}{!}{
        \begin{tabular}{cc|ccc|ccc|c}
            \label{tab:abl_gen_model}
            \multirow{2}{*}{Model} & \multirow{2}{*}{LoRA}
            & \multicolumn{3}{c|}{15-5 (ov.)}
            & \multicolumn{3}{c|}{15-1 (ov.)} & \multirow{2}{*}{MMD $\downarrow$} \\
            & & 1-15 & 16-20 & All & 1-15 & 16-20 & All & \\
            \hline
            SD2.1 & \cmark & \textbf{78.7} & \textbf{49.2} & \textbf{71.4} & \textbf{71.9} & \textbf{25.9} & \textbf{60.4} & \textbf{0.01} \\
            SD2.1 & \xmark & \textbf{78.7} & \textbf{49.2} & \underline{71.3} & \underline{66.7} & \underline{21.9} & \underline{55.5} & 0.03 \\
            SDXL & \cmark & {78.5} & 48.7 & 71.1 & 59.3 & 11.8 & 47.4 & \underline{0.02} \\
            SDXL & \xmark & 78.4 & 48.4 & 70.9 & 54.1 & 11.9 & 43.5 & 0.05 \\
        \end{tabular}
        }
    \end{minipage}\hfill
    \begin{minipage}[t]{.265\textwidth}
        \caption{Images with missing classes after 1 to 5 generations (in $\%$).}
        \label{tab:resample}
        \centering
        \setlength{\tabcolsep}{1pt}
        \renewcommand{\arraystretch}{1.1}
        \resizebox{\linewidth}{!}{
        \begin{tabular}{c|ccccc}
            Task & 1 & 2 & 3 & 4 & 5 \\
            \hline
            C2V & 12.58 & 8.29 & 6.59 & 5.55 & 5.0 \\
            15-5 (ov.) & 6.07 & 2.52 & 1.56 & 1.14 & 0.85 \\
            15-1 (ov.) & 6.55 & 2.75 & 1.54 & 1.14 & 0.9 \\
        \end{tabular}}
    \end{minipage}
\end{table*}

Following previous works, we also evaluate DR.WILSS  on the C2V benchmark~\cite{wilson}.
Here, a model is initially trained on 60 non-overlapping COCO classes (\ie, those not present in the Pascal dataset), followed by an incremental step where the remaining 20 shared classes are learned on VOC.
A key difference with the previous experiments is that the accuracy is measured on both domains.
Results in Table \ref{tab:cocovoc} show that our approach achieves very good performances effectively mitigating forgetting; in particular, the version without LoRA adapters performs  best on the  COCO dataset and ranks second on VOC data.
Here, the use of the LoRA adapters leads to a small decrease in performances (on average $0.3\%$), probably since the larger number of classes in the initial step makes~it~harder to fit all knowledge into a single adapter. Also,  the LoRAs here are trained only on COCO data and thus match its distribution.
Finally, note that unlike supervised continual SS, the ADE20K benchmark does not fit within the WILSS setting, as the non-object-centric nature of its images precludes effective image-level labeling, which is why it is generally not used.

\section{Ablation Studies} \label{ssec:ablation_study}
\noindent

\noindent \textbf{Replay Ratio:} We examine the impact of the amount of replay data per batch (interleaving rate) by repeating the 15-5 and 15-1 experiments using different values of the $\rho$ parameter. More specifically, $\rho \coloneqq \frac{n_\text{synth}}{n_\text{synth} + n_\text{real}}$, where $n_\text{synth} + n_\text{real}$ is the total batch size.
Results in Table~\ref{tab:abl_rho} show how our choice ($\rho=0.5$) consistently leads to the highest performance across the two tasks, with the only exception being the initial step of the 15-5 setting, where it is outperformed by $\rho=0.75$.
Note that generating more replay samples increases the computational cost, further supporting $\rho=0.5$ as the optimal trade-off.

\noindent \textbf{Different Diffusion Models:} We also evaluate whether using a more complex diffusion model could improve performance. Detailed results are reported in Table~\ref{tab:abl_gen_model}.
Our experiments show that replacing SD2.1 with SDXL leads to a decrease in mIoU, particularly evident in the 15-1 setting. These results are also confirmed by Maximum Mean Discrepancy (MMD) analysis.
We hypothesize that, since the semantic content is more relevant for this task than appealing visual quality, models tailored for the latter do not necessarily lead to better performance.
Concerning the inclusion of LoRA modules, it is possible to observe that their impact is limited on the single task 15-5 setting, but when multiple incremental steps are performed, as in the 15-1 setting, their impact is relevant (almost $5\%$ with SD2.1).
Additionally, we compute the cosine similarity between DINO and CLIP embeddings extracted from generated and real data samples in the 15-5 \textit{overlapped} setting. Our LoRAs achieve a relative increase of $11\%$.

As a final note, we highlight how DR.WILSS achieves SOTA performance in the 15-x tasks, even when no LoRA modules are added to the SD2.1 generative model (\eg, $+0.3\%$ mIoU on 15-5 w.r.t. the best competitor Teddy~\cite{teddy}).

\noindent \textbf{Resampling:}  
Even if diffusion models are extremely effective, ``imperfect'' generations can happen and affect the training process.
Tab.~\ref{tab:resample} shows how resampling (generating images multiple times) allows to strongly reduce the number of synthetic samples with missing target classes. 
As also discussed in Sec.~\ref{ssec:generative_module}, on the Pascal setting, the error rate is reduced below $1\%$. On the C2V task, the images with missing classes are an even more relevant issue, and again, their number can be strongly reduced by resampling. Limiting generations to 5 is a compromise between quality and computational cost.

\noindent \textbf{Statistical Relevance:} 
To estimate the significance of our results, we have re-run the 15-5 and 15-1 \textit{Ov.} experiments with 5 additional random seeds, leading to a standard deviation across all 6 seeds of $0.3$ (15-5) and $0.8$ (15-1).

\noindent \textbf{Computational Cost:}
We measure computational overhead by timing each sub-step in the 15-5 \textit{overlapped} setting. 
Training the LoRA for 50 epochs
takes 30m. 
Generating the set of 15 initial classes, with no filtering, takes 18h (9568 images at 6.7 s/img) in fp32. Note that fast generation is not the goal of our work; common optimizations can be applied without performance impact. Training the segmentation model with $\rho = 0.5$ doubles the cost (from 40m without replay to 80m).

\section{Conclusion} \label{sec:conclusion}
In this paper, we presented DR.WILSS, a novel approach to replay-based weakly supervised class-incremental semantic segmentation. 
We show that diffusion models are effective sources for replay data in weakly supervised contexts, achieving state-of-the-art results in the WILSS task across multiple benchmarks. 
Our experiments also highlight that bigger models (\ie, SDXL), whose output tends to be better appreciated by humans, are not necessarily better than smaller alternatives (\ie, SD2.1) when the synthetic images need to be processed by other architectures.
Future work will focus on improving the generative replay strategy and evaluating other diffusion models to both enhance generalization performance and handle more complex scenes, as well as analyzing the discrepancy between human preference and model performance.

\bibliographystyle{IEEEbib}
\bibliography{refs}

\end{document}